\documentclass[conference]{IEEEtran}
\IEEEoverridecommandlockouts

\usepackage{cite}
\usepackage{amsmath,amssymb,amsfonts}
\usepackage{algorithmic}
\usepackage{graphicx}
\usepackage{textcomp}
\usepackage{xcolor}
\usepackage{booktabs}
\usepackage{tabularx}
\usepackage{array}
\usepackage{url}
\usepackage[nolist]{acronym}
\def\BibTeX{{\rm B\kern-.05em{\sc i\kern-.025em b}\kern-.08em
    T\kern-.1667em\lower.7ex\hbox{E}\kern-.125emX}}
\newcommand{\FundingNotice}{%
	\begin{tabular}[t]{@{}p{0.13\columnwidth}@{\hspace{0.55em}}p{0.79\columnwidth}@{}}
		\vspace{0pt}\includegraphics[width=\linewidth]{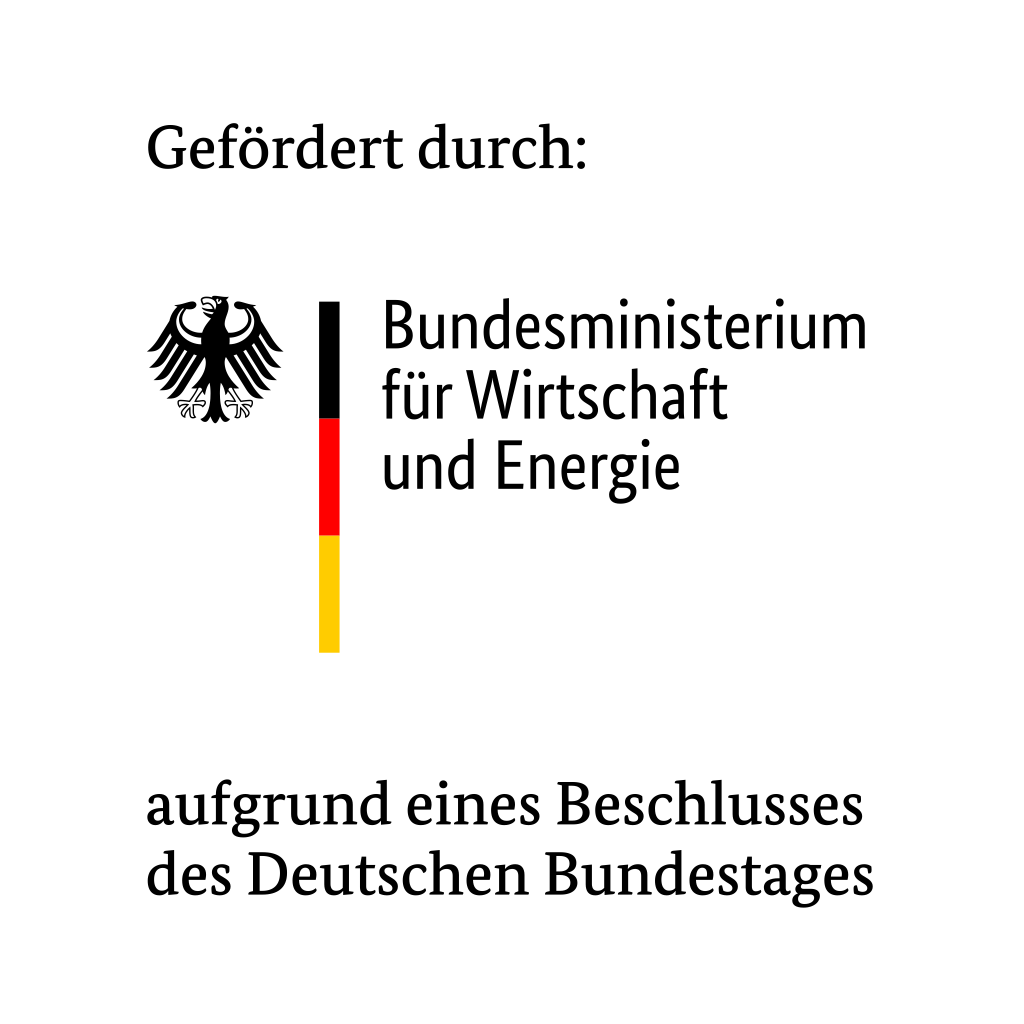} &
		\vspace{0pt}Supported by the Federal Republic of Germany, represented by the Federal Ministry for Economic Affairs and Climate Action on the basis of a decision by the German Bundestag. Funding code: 20D2401G.
	\end{tabular}%
}
\begin{document}

\begin{acronym}
	\acro{SADUSI}{\textbf{S}elf-supervised \textbf{A}nomaly \textbf{D}etection in \textbf{U}ltra\textbf{S}ound \textbf{I}maging}
\end{acronym}
\title{A Multi-Source Ultrasound Benchmark Revealing the Limits of Contemporary Self-Supervised Anomaly Detection Methods\thanks{\protect\FundingNotice}}

\author{
	\IEEEauthorblockN{Marco Riedenauer}
	\IEEEauthorblockA{\textit{University of Augsburg} \\
	marco.riedenauer@uni-a.de}
	\and
	\IEEEauthorblockN{Daniel Kienzle}
	\IEEEauthorblockA{\textit{University of Augsburg} \\
		daniel.kienzle@uni-a.de}
	\and
	\IEEEauthorblockN{Pratik Mayekar}
	\IEEEauthorblockA{\textit{University of Augsburg} \\
		pratik.mayekar@uni-a.de}
	\and
	\IEEEauthorblockN{Rainer Lienhart}
	\IEEEauthorblockA{\textit{University of Augsburg} \\
	rainer.lienhart@uni-a.de}
}

\maketitle

\begin{abstract}
Self-supervised anomaly detection is a promising paradigm for medical ultrasound, as normal images are often easier to obtain than exhaustive annotations of all possible pathologies.
However, most existing evaluations are limited to a single anatomy or task, making it unclear whether models learn a robust notion of normal ultrasound appearance or only a source-specific representation.
We introduce the \ac{SADUSI} benchmark, a multi-source ultrasound dataset designed to train and evaluate anomaly detection methods across a broad range of anatomical regions, views, and acquisition protocols.
The goal of \ac{SADUSI} is to provide a diverse normal ultrasound distribution and a benchmark for visible structural anomalies that can be assessed from single images.
We evaluate representative self-supervised anomaly detection methods and find that current approaches struggle in this setting.
In particular, reconstruction-based diffusion methods such as AnoDDPM and DeCo-Diff achieve pixel-level AUROC values of 0.56--0.72 and maximum F1 scores of 0.10--0.26, indicating limited separation of pathology from normal image regions.
Feature-based PatchCore variants perform better, reaching pixel-level AUROC values of 0.76--0.83, but remain limited with maximum F1 scores of 0.14--0.40.
These findings suggest that broad multi-source ultrasound anomaly detection remains an open challenge and that \ac{SADUSI} can serve as a resource for developing methods that generalize beyond anatomy-specific settings.
\end{abstract}


\section{Introduction}
Anomaly detection aims to identify observations that deviate from an expected notion of normality \cite{ruff2021unifying}.
Statistically, anomalies are rare or unlikely samples under the data distribution observed during training.
In visual inspection, only a subset of such statistical deviations is domain-relevant: medical anomalies may correspond to visible pathologies such as tumors, nodules, or deposits, while industrial anomalies may correspond to material defects such as cracks, inclusions, or voids \cite{schlegl2017unsupervised,bergmann2019mvtec,wang2022ultrasonic}.

Medical ultrasound imaging is widely used because it is comparatively inexpensive, real-time capable, portable, and free of ionizing radiation.
Automatic detection of suspicious structures in ultrasound images could support clinical workflows by highlighting potentially relevant findings and reducing repetitive screening effort.
Fully supervised training, however, remains difficult because relevant pathologies are diverse and often rare, pixel-level annotations require clinical expertise, and ultrasound appearance varies strongly across anatomy, view, device, operator, and acquisition protocol.
Self-supervised and unsupervised anomaly detection methods address part of this limitation by learning from visually normal data and identifying deviations at inference time \cite{schlegl2017unsupervised,huang2023selfsupervised}.

Most existing medical ultrasound anomaly detection settings are anatomy- or task-specific.
Models trained on one region may learn organ- or view-specific appearance rather than a broader notion of normal ultrasound that transfers across fetal, thyroid, liver, breast, lung, cardiac, or abdominal imaging.
Conversely, a broadly trained model must distinguish visible pathological structures from the large variability of normal anatomy.

This work focuses on the visual inspection setting.
We do not aim to build a dataset that contains all possible pathologies, nor do we assume that every clinically relevant abnormality can be detected from a single ultrasound image.
Functional or temporal abnormalities may require video, measurements, or broader clinical context and are therefore outside the scope of this single-image benchmark.
Instead, our goal is to provide a broad multi-source ultrasound dataset that supports self-supervised learning of heterogeneous normal ultrasound appearance and enables evaluation on visible structural anomalies where suitable masks are available.

Despite the relevance of this setting, there is currently no large-scale public dataset specifically designed for training and evaluating self-supervised anomaly detection methods across multiple medical ultrasound domains.
Existing anomaly detection benchmarks have advanced industrial visual inspection \cite{bergmann2019mvtec}, but they do not capture anatomical variability, acquisition-dependent appearance, speckle patterns, and clinical label limitations of ultrasound imaging, while many medical anomaly detection studies rely on private or task-specific datasets.
This gap makes it difficult to assess whether methods learn robust ultrasound representations or merely exploit dataset-specific appearance patterns.

To address this limitation, we introduce the \ac{SADUSI} benchmark, a multi-source ultrasound dataset for training and evaluating self-supervised anomaly detection methods.\footnote{Repository: \url{https://github.com/MarcoRiedenauer/SADUSI}}
It combines public ultrasound sources from different anatomical regions, acquisition protocols, and image formats for visible structural anomaly detection in single frames.
Our benchmark shows that current methods struggle in this broad setting: reconstruction-based diffusion methods show limited pathology localization, while feature-based methods such as PatchCore perform better but remain limited.

To summarize, the main contributions of this paper are:
\begin{itemize}
	\item We curate \ac{SADUSI}, a broad multi-source ultrasound dataset for self-supervised anomaly detection.
	\item We benchmark representative reconstruction, feature-memory, and student-teacher anomaly detection methods.
	\item We show that source-domain variation can dominate pathology-related variation, limiting current methods in this setting.
\end{itemize}

\section{Related Work}
Public ultrasound datasets have enabled substantial progress in supervised medical image analysis, including fetal measurement and plane recognition~\cite{vandenheuvel2018hc18,burgos2020fetalplanes}, thyroid and breast lesion analysis~\cite{gong2023tn3k,al2020busi,vallez2025bus_uclm}, echocardiography~\cite{ouyang2020echonet,leclerc2019camus}, and lung ultrasound for COVID-19 analytics~\cite{ebadi2021covidx_us}.
These datasets are valuable resources for segmentation, classification, measurement, or disease-specific prediction tasks, but they are usually released and evaluated in isolation.
Consequently, they do not directly provide a unified benchmark for training self-supervised anomaly detection methods on heterogeneous normal ultrasound appearance and evaluating visible structural anomalies across multiple anatomical domains.

Deep learning in medical imaging often relies on task-specific labels, architectures, and pretraining strategies.
Self-supervised learning has therefore become increasingly important because it can exploit large unlabeled or weakly labeled medical image collections~\cite{huang2023selfsupervised}.
For example, ~\cite{kienzle2022covid_convnext} study custom pretraining strategies for COVID detection and severity prediction in 3D CT data.
Such work highlights the importance of domain-adapted representation learning in medical imaging, but it still addresses a supervised, disease-specific setting rather than broad anomaly detection from normal training data.

Anomaly detection has been studied through classical one-class, probabilistic, reconstruction-based, and deep representation-based formulations~\cite{ruff2021unifying}.
In medical imaging, reconstruction-based approaches such as GAN-based anomaly detection have been used to model normal anatomy and identify deviations~\cite{schlegl2017unsupervised}.
In visual inspection, benchmarks such as MVTec AD have driven feature-memory and student-teacher methods, including PatchCore and EfficientAD~\cite{bergmann2019mvtec,roth2022patchcore,batzner2024efficientad}.
More recently, diffusion-based methods such as AnoDDPM and DeCo-Diff have reformulated anomaly detection as reconstruction or correction of deviations from normality~\cite{wyatt2022anoddpm,beizaee2025decodiff}.
However, these methods are commonly evaluated either on industrial data or on comparatively narrow medical settings.
\ac{SADUSI} addresses this gap by combining diverse ultrasound sources into a multi-source benchmark for visible structural anomaly localization.

\section{Dataset}

\subsection{Task Definition and Scope}
\label{sec:task}
\ac{SADUSI} is designed as a multi-source dataset for training and evaluating single-image anomaly detection methods on medical ultrasound images.
Its primary purpose is to expose self-supervised models to a broad normal ultrasound distribution.
In this work, the term \textit{anomaly} strictly refers to a visible structural anomaly, i.e., a pathological finding that can be identified in a single ultrasound image.
This excludes abnormalities that require temporal, functional, or broader clinical context.
For example, a cardiac rhythm disorder would be treated as normal if the corresponding single echocardiography frame does not show a visible structural abnormality.

Let $\mathcal{X} \subseteq \mathbb{R}^{h \times w \times c}$ denote the space of ultrasound images.
For the structural anomaly detection setting, we represent the annotated dataset as
\begin{equation}
	\mathcal{D} = \{(x_i, y_i, m_i, z_i)\}_{i=1}^{N},
\end{equation}
where $s_i=(x_i,y_i,m_i,z_i)$ denotes one dataset entry, $x_i \in \mathcal{X}$ is an ultrasound image, $y_i \in \{0,1\}$ is an image-level binary structural label, $m_i$ is an optional pixel-level mask, and $z_i$ summarizes source metadata such as anatomy, view, acquisition protocol, and source dataset.
The label $y_i=0$ denotes, according to the available source labels and metadata, an image that is considered visually normal, while $y_i=1$ denotes an image with a visible structural anomaly.
Examples of structural anomalies include nodules, tumors, deposits, or other localized morphology changes that are visible in the image itself.
The corresponding normal and anomalous subsets are defined as
\begin{equation}
	\mathcal{D}_n = \{s_i \in \mathcal{D} \mid y_i = 0\},\quad
	\mathcal{D}_a = \{s_i \in \mathcal{D} \mid y_i = 1\}.
\end{equation}
If pixel-level annotations are available, the corresponding localization ground truth is given by
\begin{equation}
	\mathcal{G}_a = \{(x_i,m_i) \mid s_i \in \mathcal{D}_a,\, m_i \neq \emptyset\}.
\end{equation}
For datasets without pixel-level annotations, $m_i$ is undefined.
Such datasets could support image-level anomaly detection based on $y_i$, but they are not used for the segmentation benchmark in this work.

\subsection{Source Pool and Harmonization}
\ac{SADUSI} aggregates public ultrasound image datasets from multiple anatomical regions, acquisition protocols, and original data formats.
Let $\mathcal{S}=\{S_k\}_{k=1}^{K}$ denote the set of source datasets, where each source contributes a subset $\mathcal{D}^{(k)}$ and the converted pool is $\mathcal{D}=\bigcup_{k=1}^{K}\mathcal{D}^{(k)}$.
The dataset is designed to cover variation factors that affect ultrasound imaging in practice, including anatomy, view, acquisition protocol, device characteristics, operator-dependent appearance, resolution, field of view, and image quality.
This heterogeneity is central to the intended self-supervised setting where a model trained on \ac{SADUSI} should not only learn normality for one specific anatomy, but should be exposed to the diversity of normal ultrasound appearance across many regions and acquisition settings.
By combining diverse sources, \ac{SADUSI} aims to support the development and evaluation of anomaly detection methods that generalize beyond very narrow-domain anatomy-specific benchmarks.

All source datasets are converted into a unified image-based representation.
Raw source items are mapped to one or more 2D images: still-image datasets are imported directly, videos, 3D ultrasound sources, and cine clips are converted by extracting every fifth frame.
This sampling strategy reduces the strong temporal redundancy of ultrasound sequences while preserving representative variations in probe motion, anatomy, and image appearance.

After conversion, each image is manually inspected and harmonized to reduce source-specific artifacts that are unrelated to the ultrasound content.
In particular, black borders are cropped where possible, and visible device user interfaces, mouse cursors, measurement overlays, or other non-anatomical markings are removed if they appear in the image area.
In Section \ref{sec:task}, $x_i$ refers to the harmonized image.
This step is important because anomaly detection methods may otherwise learn shortcuts from dataset-specific layouts or annotations instead of modeling anatomical image content.
During conversion, source-specific metadata and labels are preserved where available and linked to the corresponding extracted images.

\subsection{Filtered Normal Subset}
The larger converted normal pool $\mathcal{D}_n$ may contain implicit anomalies.
Many public ultrasound datasets are released for tasks such as organ segmentation, plane classification, or measurement, and their documentation does not always state whether visible pathologies are absent.
If such images are used as normal training data, a self-supervised model may learn pathological structures as part of the normal distribution and lose sensitivity to the masked pathologies used for evaluation.
We therefore define a filtered normal subset $\mathcal{D}^{\mathrm{f}}_n \subseteq \mathcal{D}_n$ as a conservative alternative to the full converted normal pool.
It only contains normal training sources for which the source labels, metadata, or dataset description explicitly support the absence of visible structural anomalies.
Sources with explicitly labeled structural abnormalities remain part of the anomalous pool $\mathcal{D}_a$. 
No separate filtered anomalous subset is introduced.
This filtering cannot guarantee that every normal image is clinically healthy, since visible pathologies may be missed during clinical assessment or annotation.
Thus, normality is defined operationally from the available annotations and documentation.
We report results for both, the full converted normal pool $\mathcal{D}_n$ and the filtered normal subset $\mathcal{D}^{\mathrm{f}}_n$: the former reflects the amount of visually plausible normal ultrasound data available from public sources, while the latter is the conservative training pool used when normal status must be explicitly supported.
The filtered normal subset is defined as
\begin{equation}
	\mathcal{D}^{\mathrm{f}}_n = \{s_i \in \mathcal{D}_n \mid \text{normal status is explicitly supported}\}.
\end{equation}

Table~\ref{tab:source_datasets} documents the converted source pool and its contribution to the filtered normal subset.

\begin{table}[t]
	\caption{Source datasets included in \ac{SADUSI}.}
	\label{tab:source_datasets}
	\centering
	\tiny
	\setlength{\tabcolsep}{2pt}
	\renewcommand{\arraystretch}{0.95}
	\begin{tabularx}{\columnwidth}{>{\raggedright\arraybackslash}p{0.35\columnwidth}
			>{\raggedright\arraybackslash}p{0.16\columnwidth}
			>{\raggedright\arraybackslash}p{0.10\columnwidth}
			>{\centering\arraybackslash}p{0.22\columnwidth}
			>{\centering\arraybackslash}p{0.10\columnwidth}}
		\toprule
		Source & Anatomy & Type & $|\mathcal{D}_n|$ / $|\mathcal{D}^{\mathrm{f}}_n|$ / $|\mathcal{D}_a|$ & Use \\
		\midrule
		AbdomenUS~\cite{orlando_ussimsegm} & Abdominal & Images & 616 / 616 / 0 & Train \\
		AUL~\cite{annotated_ultrasound_liver} & Liver & Images & 100 / 100 / 434 & Train / Test \\
		Breast-Lesions-USG~\cite{pawlowska2024breast,tcia_breast_lesions_usg} & Breast & Images & 4 / 4 / 0 & Train \\
		BUS-UCLM~\cite{vallez2024bus_uclm,vallez2025bus_uclm} & Breast & Images & 405 / 405 / 0 & Train \\
		BUSI~\cite{al2020busi,busi_kaggle} & Breast & Images & 131 / 131 / 512 & Train / Test \\
		COVID-19~\cite{sedik2020covid_ultrasound} & Lung & Images & 62 / 62 / 0 & Train \\
		EchoNetPediatric~\cite{reddy2023echonet_peds,echonet_peds_dataset} & Echocardiography & Videos & 159,006 / 0 / 0 & Train \\
		FALLMUD~\cite{fallmud_dataset} & Muscle & Images & 813 / 0 / 0 & Train \\
		FASS~\cite{fetal_abdominal_dataset_2023} & Fetal abdominal & Images & 1,577 / 1,577 / 0 & Train \\
		Fast-U-Net~\cite{ashkani2022fastunet} & Fetal & Images & 79 / 0 / 0 & Train \\
		HC18~\cite{vandenheuvel2018hc18,hc18_dataset} & Fetal head & Images & 1,333 / 1,333 / 0 & Train \\
		NFLD~\cite{hoque2026nafld} & Liver & Images & 11 / 11 / 0 & Train \\
		SegThy~\cite{segthy_dataset} & Thyroid & 3D volumes & 1,962 / 1,962 / 0 & Train \\
		STMUS\_NDA~\cite{marzola2021muscle,marzola2021dataset} & Muscle & Images & 6,749 / 0 / 0 & Train \\
		TG3K~\cite{gong2023thyroid} & Thyroid & Images & 3,585 / 0 / 0 & Train \\
		TN3K~\cite{gong2023tn3k} & Thyroid & Images & 0 / 0 / 1,349 & Test \\
		UBPD~\cite{ding2022mallesnet} & Brachial plexus & Images & 955 / 0 / 0 & Train \\
		Ultrasound-Fetus~\cite{anitha2024fetus} & Fetal & Images & 335 / 335 / 0 & Train \\
		Ultrasound-OpticNerveSheaths~\cite{marzola2023optic_nerve,marzola2023optic_dataset} & Optic nerve & Images & 464 / 0 / 0 & Train \\
		\midrule
		Total & Multi-anatomy & Mixed & 178,187 / 6,536 / 2,295 & -- \\
		\bottomrule
	\end{tabularx}
	\vspace{-3ex}
\end{table}

The filtered normal subset combines fetal, thyroid and neck, liver, abdominal, breast, and lung ultrasound images.
The anomalous evaluation pool currently includes visible structural findings from liver, breast, and thyroid ultrasound datasets, including focal liver masses, breast pathologies, and thyroid nodules.
The current benchmark uses only anomalous datasets with segmentation masks and evaluates pixel-level anomaly localization. 
Datasets without masks are not used for the current evaluation.

\section{Benchmark Setup}
We benchmark \ac{SADUSI} on approaches for pixel-level anomaly segmentation.
The benchmark evaluates whether methods trained on a heterogeneous set of ultrasound images considered visually normal can localize visible structural anomalies in held-out anomalous images.
Our benchmark includes four representative self-supervised anomaly detection methods: DeCo-Diff~\cite{beizaee2025decodiff}, AnoDDPM~\cite{wyatt2022anoddpm}, PatchCore~\cite{roth2022patchcore}, and EfficientAD~\cite{batzner2024efficientad}.
DeCo-Diff and AnoDDPM represent diffusion-based reconstruction methods, PatchCore represents feature-embedding methods based on nearest-neighbor matching in a memory bank, and EfficientAD represents efficient student-teacher anomaly detection with an additional autoencoder component.
For PatchCore, we evaluate two backbone variants: a WideResNet-50 feature extractor (PatchCore-WRN) based on Wide Residual Networks~\cite{zagoruyko2016wide}, and a DINOv3 ViT-B/16 feature extractor (PatchCore-DINOv3)~\cite{simeoni2025dinov3}.
All methods produce pixel-level anomaly maps and can therefore be evaluated on the segmentation masks available in $\mathcal{G}_a$.
A model is trained on
\begin{equation}
	\mathcal{D}_{\mathrm{train}} \subseteq \mathcal{D}_n
\end{equation}
and learns an anomaly scoring function $s_\theta: \mathcal{X} \rightarrow \mathbb{R}^{h \times w}$.
At inference time, $s_\theta(x)$ produces a pixel-level anomaly map for an input image $x$.

\subsection{Training Protocol}
Each method is trained self-supervised on either the full normal pool $\mathcal{D}_n$ or the filtered pool $\mathcal{D}^{\mathrm{f}}_n$, which contains only sources whose visually normal status is explicitly supported.
Both variants use the same masked anomalous evaluation set $\mathcal{G}_a$ to isolate the effect of training-pool composition.

All runs use seed 1000 and no validation set for model selection.
AnoDDPM, DeCo-Diff, and EfficientAD use 300k/11,341/70k steps with effective batch sizes 1/256/4, respectively. 
PatchCore performs one batch-8 feature pass and retains 10\% of the extracted training patch features after coreset subsampling, while EfficientAD reserves 10\% of its pool only to estimate anomaly-map normalization statistics.
Since PatchCore-DINOv3 gives the strongest AUROC in the initial comparison, we additionally evaluate an augmented PatchCore-DINOv3 variant, denoted by ``(A)'', to test whether mild acquisition-like variability in the support memory improves localization.

\subsection{Preprocessing}
All images and masks are converted to a spatial size of $256 \times 256$ pixels before training and inference.
To avoid distorting anatomical structures, images are resized while preserving the original aspect ratio and padded to the target resolution.
The same spatial transformation is applied to segmentation masks using nearest-neighbor interpolation.

Apart from this dataset-level harmonization, we keep the method-specific input preprocessing consistent with the respective original setups~\cite{beizaee2025decodiff,wyatt2022anoddpm,roth2022patchcore,batzner2024efficientad}.
For RGB-based methods, grayscale ultrasound images are converted to three channels by duplicating the single intensity channel.
Both PatchCore variants use ImageNet-normalized inputs for their pretrained WRN and DINOv3 feature extractors, and EfficientAD applies the corresponding normalization internally in its teacher, student, and autoencoder networks.
The diffusion-based reconstruction methods use their original intensity scaling: DeCo-Diff operates on images normalized to $[-1,1]$ for the latent diffusion pipeline, while AnoDDPM is trained and evaluated in a single-channel setting.
Except for the explicit PatchCore-DINOv3 augmentation experiment, no additional cropping, geometric, intensity, or synthetic augmentations are applied beyond dataset conversion and harmonization.
For this variant, PatchCore stores each original training image plus one augmented support view generated by mild ultrasound-preserving transformations: affine perturbations ($\pm7^\circ$, $\pm4\%$ translation, $0.95$--$1.05$ scale), horizontal flips, CLAHE, gamma changes in $[0.85,1.15]$, brightness/contrast changes up to $0.1$, and multiplicative speckle noise with $\sigma=0.03$.

\subsection{Evaluation Metrics}
The benchmark evaluates pixel-level anomaly localization.
For each test image, a method predicts an anomaly map $\hat{m}_i \in [0,1]^{H \times W}$, which is compared against the corresponding binary ground-truth mask $m_i$.
We report pixel-level AUROC, average precision (AP), AUPRO, maximum F1 score ($F1_{\max}$), precision, and recall.
AUPRO measures the area under the per-region overlap curve and summarizes how well predicted anomaly regions overlap with connected ground-truth anomaly regions across thresholds.

Threshold-independent metrics are computed directly from the continuous anomaly maps.
For threshold-dependent metrics such as precision, recall, and F1, we sweep thresholds over the anomaly maps and select the threshold $\tau^\ast = \arg\max_{\tau} F1(\tau)$ to compute their values.
Thus, $F1_{\max}$, precision, and recall measure the best achievable operating point on the evaluated segmentation masks, while AUROC, AP, and AUPRO summarize ranking quality without fixing a threshold.

\section{Experiments and Results}
In Table \ref{tab:results}, we report pixel-level anomaly localization results on three anomalous ultrasound subsets with segmentation masks for thyroid nodule, breast, and liver ultrasound evaluations.

\begin{table}[t]
	\caption{Pixel-level anomaly localization results. ``(A)'' denotes data augmentation.}
	\label{tab:results}
	\centering
	\scriptsize
	\renewcommand{\arraystretch}{0.94}
	\resizebox{\columnwidth}{!}{%
	\begin{tabular}{lllcccccc}
		\toprule
		Eval. set & Training & Method & AUROC & AP & AUPRO & F1 & Prec. & Rec. \\
		\midrule
		TN3K & Filtered & AnoDDPM & 0.59 & 0.14 & 0.17 & 0.23 & 0.14 & 0.82 \\
		TN3K & Filtered & DeCo-Diff & 0.66 & 0.16 & 0.20 & 0.26 & 0.16 & 0.64 \\
		TN3K & Filtered & PatchCore-WRN & 0.76 & 0.24 & \textbf{0.40} & 0.33 & 0.22 & 0.62 \\
		TN3K & Filtered & PatchCore-DINOv3 & \textbf{0.79} & \textbf{0.26} & 0.29 & \textbf{0.36} & \textbf{0.26} & 0.61 \\
		TN3K & Filtered & PatchCore-DINOv3 (A) & 0.76 & 0.24 & 0.37 & 0.33 & 0.22 & 0.62 \\
		TN3K & Filtered & EfficientAD & 0.50 & 0.11 & 0.07 & 0.20 & 0.11 & \textbf{0.94} \\
		\midrule
		TN3K & Full & AnoDDPM & 0.61 & 0.15 & 0.19 & 0.24 & 0.14 & 0.84 \\
		TN3K & Full & DeCo-Diff & 0.60 & 0.13 & 0.15 & 0.24 & 0.14 & 0.87 \\
		TN3K & Full & PatchCore-WRN & 0.77 & 0.26 & 0.42 & 0.34 & 0.25 & 0.57 \\
		TN3K & Full & PatchCore-DINOv3 & \textbf{0.82} & \textbf{0.32} & 0.34 & \textbf{0.40} & \textbf{0.31} & 0.58 \\
		TN3K & Full & PatchCore-DINOv3 (A) & 0.78 & 0.27 & \textbf{0.43} & 0.35 & 0.25 & 0.57 \\
		TN3K & Full & EfficientAD & 0.53 & 0.12 & 0.14 & 0.21 & 0.12 & \textbf{0.88} \\
		\midrule
		BUSI & Filtered & AnoDDPM & 0.62 & 0.11 & 0.23 & 0.18 & 0.11 & 0.42 \\
		BUSI & Filtered & DeCo-Diff & 0.68 & 0.14 & 0.40 & 0.21 & 0.15 & 0.38 \\
		BUSI & Filtered & PatchCore-WRN & \textbf{0.82} & \textbf{0.27} & \textbf{0.66} & \textbf{0.33} & \textbf{0.25} & 0.47 \\
		BUSI & Filtered & PatchCore-DINOv3 & \textbf{0.82} & 0.23 & 0.51 & \textbf{0.33} & 0.24 & 0.54 \\
		BUSI & Filtered & PatchCore-DINOv3 (A) & 0.81 & \textbf{0.27} & \textbf{0.66} & \textbf{0.33} & \textbf{0.25} & 0.47 \\
		BUSI & Filtered & EfficientAD & 0.50 & 0.08 & 0.30 & 0.15 & 0.08 & \textbf{0.92} \\
		\midrule
		BUSI & Full & AnoDDPM & 0.63 & 0.11 & 0.24 & 0.18 & 0.11 & 0.55 \\
		BUSI & Full & DeCo-Diff & 0.56 & 0.09 & 0.14 & 0.17 & 0.09 & \textbf{0.91} \\
		BUSI & Full & PatchCore-WRN & 0.80 & 0.23 & \textbf{0.64} & 0.30 & 0.22 & 0.46 \\
		BUSI & Full & PatchCore-DINOv3 & \textbf{0.82} & 0.22 & 0.51 & \textbf{0.33} & \textbf{0.24} & 0.56 \\
		BUSI & Full & PatchCore-DINOv3 (A) & 0.79 & \textbf{0.24} & \textbf{0.64} & 0.30 & 0.23 & 0.46 \\
		BUSI & Full & EfficientAD & 0.55 & 0.09 & 0.32 & 0.16 & 0.09 & 0.78 \\
		\midrule
		AUL & Filtered & AnoDDPM & 0.72 & 0.06 & 0.26 & 0.13 & 0.07 & 0.64 \\
		AUL & Filtered & DeCo-Diff & 0.66 & 0.05 & 0.19 & 0.10 & 0.05 & 0.88 \\
		AUL & Filtered & PatchCore-WRN & \textbf{0.79} & 0.07 & \textbf{0.41} & \textbf{0.16} & \textbf{0.09} & 0.82 \\
		AUL & Filtered & PatchCore-DINOv3 & 0.77 & \textbf{0.09} & 0.36 & 0.14 & \textbf{0.09} & 0.31 \\
		AUL & Filtered & PatchCore-DINOv3 (A) & \textbf{0.79} & 0.08 & 0.39 & \textbf{0.16} & \textbf{0.09} & 0.82 \\
		AUL & Filtered & EfficientAD & 0.34 & 0.03 & 0.07 & 0.07 & 0.04 & \textbf{1.00} \\
		\midrule
		AUL & Full & AnoDDPM & 0.72 & 0.07 & 0.27 & 0.13 & 0.07 & 0.64 \\
		AUL & Full & DeCo-Diff & 0.71 & 0.05 & 0.23 & 0.12 & 0.06 & \textbf{0.91} \\
		AUL & Full & PatchCore-WRN & 0.80 & 0.08 & 0.42 & 0.16 & 0.09 & 0.78 \\
		AUL & Full & PatchCore-DINOv3 & \textbf{0.83} & \textbf{0.11} & \textbf{0.45} & \textbf{0.18} & \textbf{0.11} & 0.47 \\
		AUL & Full & PatchCore-DINOv3 (A) & 0.81 & 0.08 & \textbf{0.45} & 0.17 & 0.09 & 0.77 \\
		AUL & Full & EfficientAD & 0.49 & 0.03 & 0.10 & 0.08 & 0.04 & 0.85 \\
		\bottomrule
	\end{tabular}%
	}
	\vspace{-4ex}
\end{table}

Overall, the PatchCore variants achieve the strongest localization results, with best AUROC values of 0.82 on TN3K and BUSI and 0.83 on AUL, but the corresponding $F1_{\max}$ values of 0.40, 0.33, and 0.18 show that pixel-level delineation remains difficult. PatchCore-WRN often yields competitive or higher AUPRO values, while augmented PatchCore-DINOv3 improves AUPRO in several settings, most clearly on TN3K and BUSI, but not the best AUROC or $F1_{\max}$ values. AnoDDPM and DeCo-Diff generally yield lower AUROC and $F1_{\max}$ values with high recall but low precision, and EfficientAD performs close to or below random ranking on several evaluations. Neither training pool is uniformly superior.

\subsection{Anomaly-Map Contrast Analysis}
The pixel-level metrics in Table~\ref{tab:results} summarize thresholded localization performance and ranking quality.
To further inspect the produced anomaly maps, we measure whether each method assigns higher anomaly scores inside the annotated structural abnormality than in the remaining image background.
For an anomalous image $x_i$ with anomaly map $A_i$ and binary ground-truth mask $M_i$, we compute the anomaly-background contrast
\begin{equation}
	\Delta_i =
	\frac{1}{|M_i|}\sum_{p \in M_i} A_i(p)
	-
	\frac{1}{|\Omega_i \setminus M_i|}\sum_{p \in \Omega_i \setminus M_i} A_i(p),
\end{equation}
where $\Omega_i$ denotes the image domain.
Positive $\Delta_i$-values indicate that the annotated pathology receives higher anomaly scores than the surrounding background.
Because anomaly-map scales differ across methods, we robustly normalize the contrast values separately for each model variant and anomaly-map type. Let $q^{(m)}_{0.025}$ and $q^{(m)}_{0.975}$ denote the corresponding quantiles over all contrast values of method $m$. We compute
\begin{equation}
	\widetilde{\Delta}_i = \operatorname{clip}_{[0,1]}\!\left(\frac{\Delta_i-q^{(m)}_{0.025}}{q^{(m)}_{0.975}-q^{(m)}_{0.025}}\right).
\end{equation}
Thus, the central 95\% of each method's contrast range defines the linear scaling interval, while values outside it are retained and clipped to 0 or 1.

\begin{figure*}[t]
	\centering
	\includegraphics[width=\textwidth]{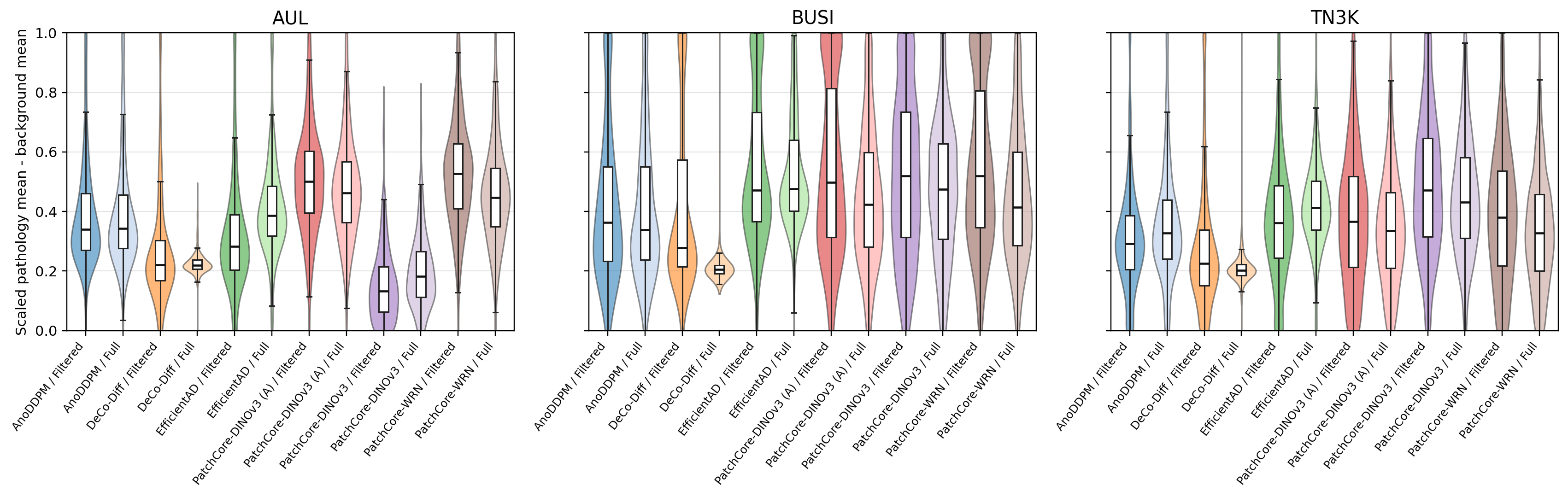}
	\vspace{-3ex}
	\caption{Scaled pathology-background anomaly-map contrast by evaluation dataset, model, and training pool. Violin plots show the distributions and embedded box plots show their medians and interquartile ranges.}
	\label{fig:model_contrast_analysis}
	\vspace{-3ex}
\end{figure*}

Figure~\ref{fig:model_contrast_analysis} compares these scaled contrast distributions by evaluation dataset and training pool.
PatchCore-DINOv3 and PatchCore-WRN show relatively high scaled contrasts on BUSI, while PatchCore-DINOv3 also performs strongly on TN3K.
On AUL, however, PatchCore-DINOv3 shifts toward the lower end of its method-specific range, whereas PatchCore-WRN shows stronger filtered-run contrast.
AnoDDPM changes little between the training pools, while DeCo-Diff trained on the full pool produces particularly narrow low-valued distributions across all datasets.
The effect of filtering is therefore model- and dataset-dependent rather than uniformly beneficial.
Since each method is scaled independently, the figure supports comparison of relative distribution shapes and training-set effects, but not absolute raw contrast magnitudes between methods.

\subsection{Feature-Space Analysis}
To better understand the limited localization performance, we analyze both normal training pools in a pretrained DINOv3 ViT-B/16 embedding space~\cite{simeoni2025dinov3}.
Images and ground-truth masks are transformed to $256 \times 256$ pixels using aspect-ratio preserving resizing and zero padding, with bilinear and nearest-neighbor interpolation, respectively.
DINOv3 produces a feature tensor in $\mathbb{R}^{768 \times 16 \times 16}$.
Only the patch tokens are retained, excluding class tokens.
Each grid position is treated as one 768-dimensional patch embedding and normalized individually using $\ell_2$ normalization.
The corresponding mask is reduced to the same $16 \times 16$ grid by average pooling, yielding the fraction of pathological pixels within each patch region.
A patch is labeled anomalous if this fraction exceeds $0.5$ and normal otherwise. 
Consequently, normal patches may originate from normal images or regions outside the annotated pathology in anomalous images.
Distances are computed as Euclidean distances between normalized patch embeddings.

For each training pool, we sample up to 500 balanced pairs per dataset pair for normal--normal and normal--anomalous comparisons, separated into within- and cross-dataset cases.
We further query each anomalous patch exhaustively against all normal patches from other images using rank-1 nearest-neighbor search and analyze its $k=10$ neighborhood among all patches, excluding the query image to prevent spatially adjacent regions of the same pathology from dominating.
This analysis is not used for training. 
It characterizes local dataset structure and whether filtering simplifies the task.

\begin{figure*}[t]
	\centering
	\includegraphics[width=0.325\textwidth]{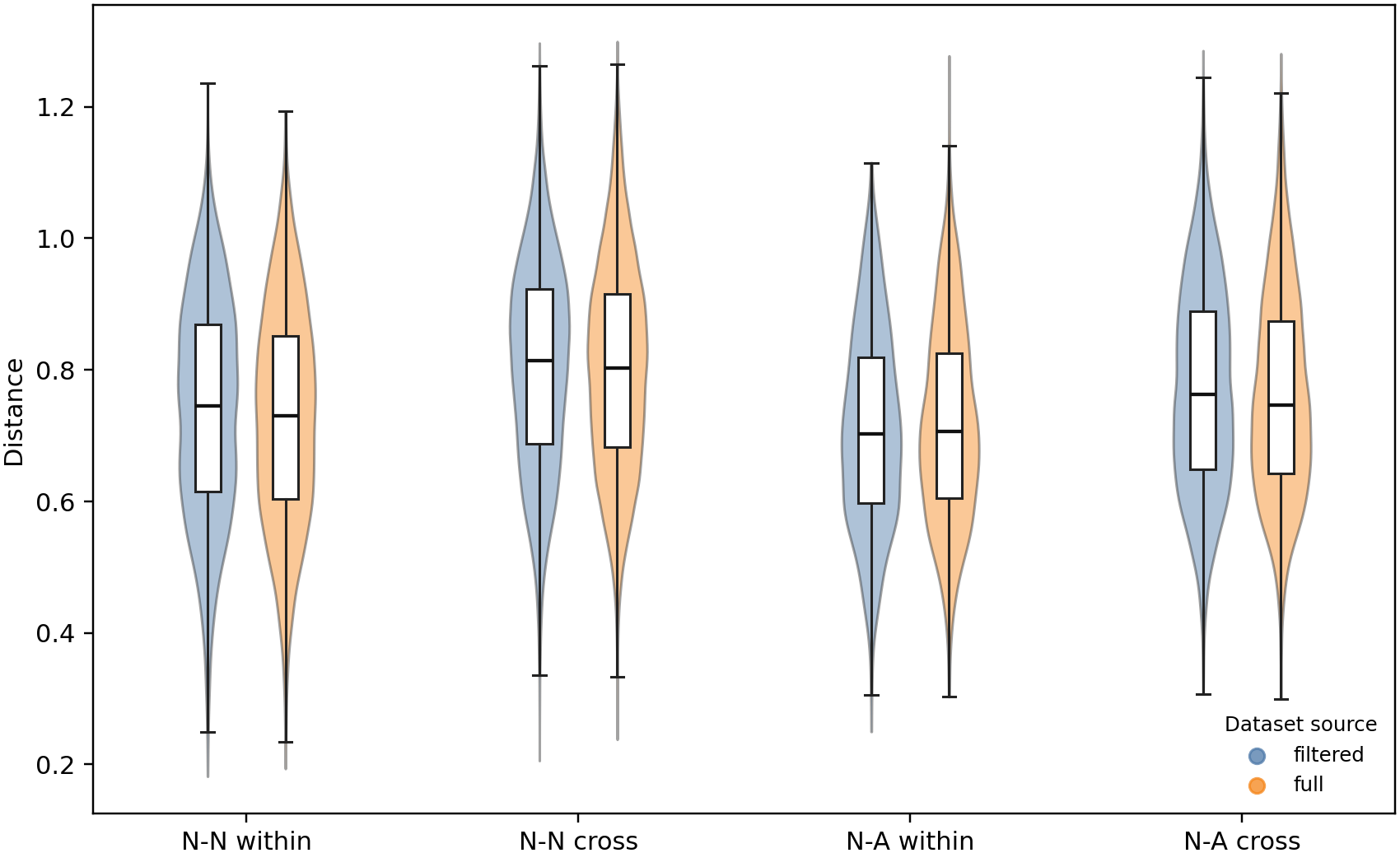}
	\hfill
	\includegraphics[width=0.325\textwidth]{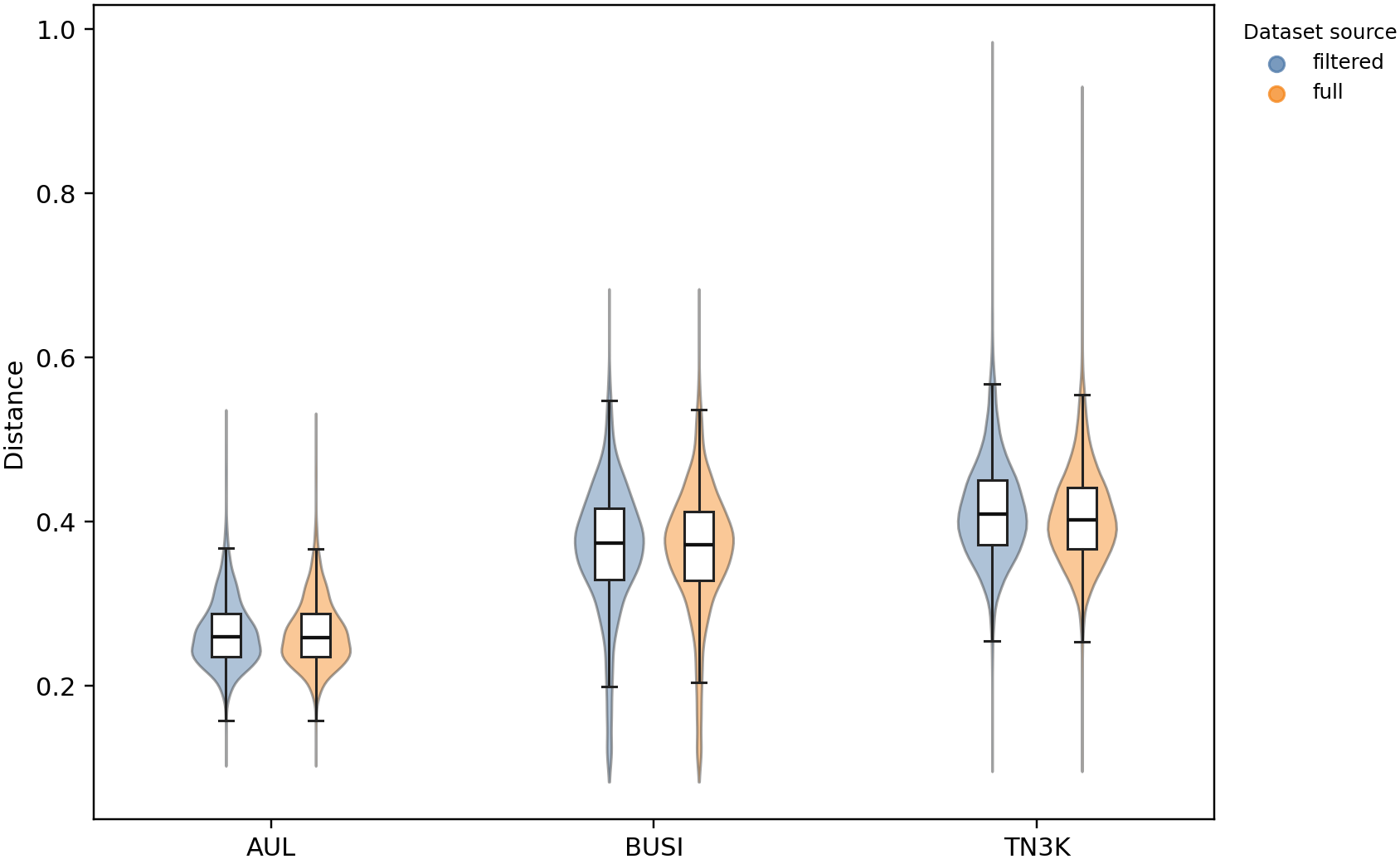}
	\hfill
	\includegraphics[width=0.325\textwidth]{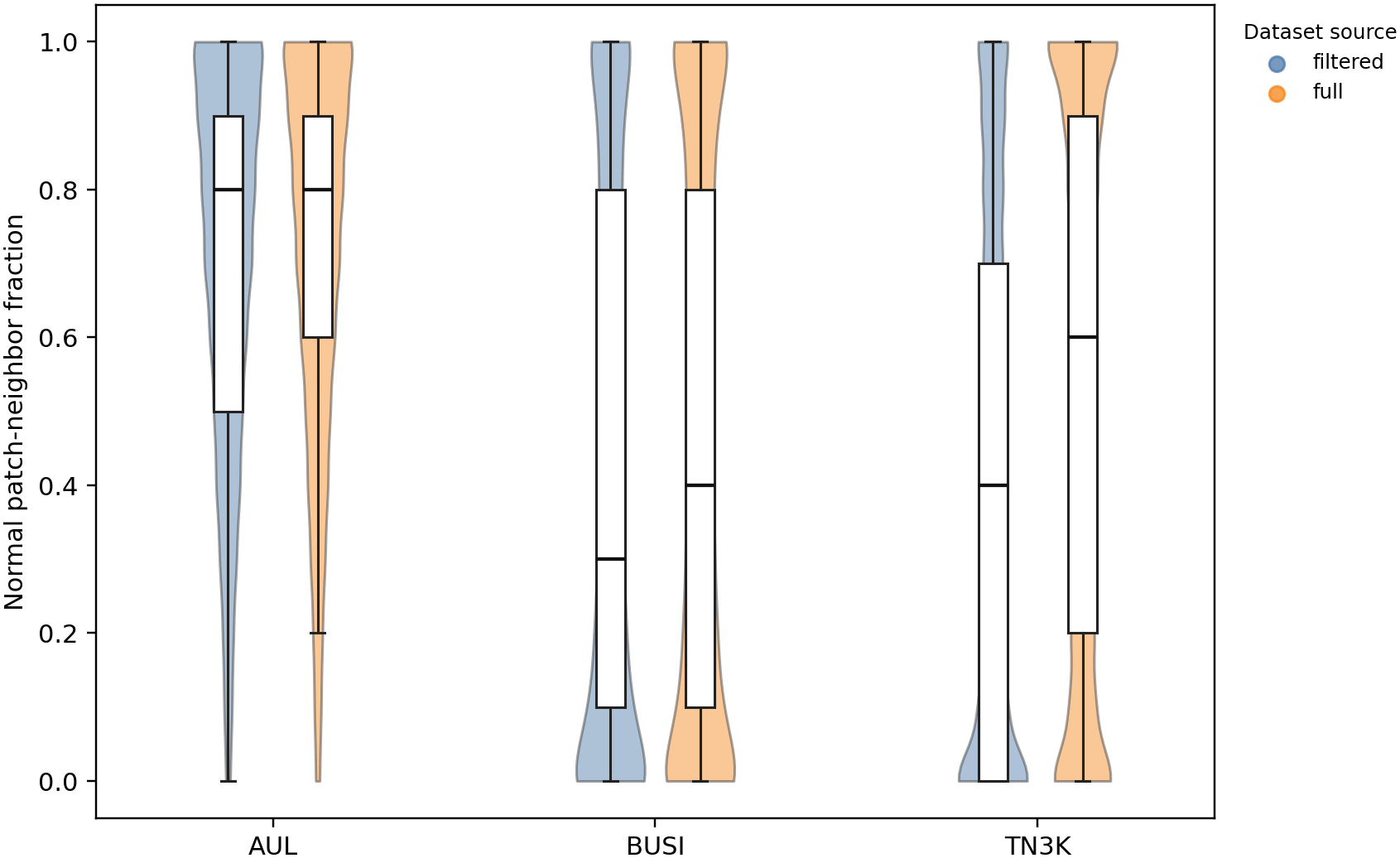}
	\vspace{-2ex}
	\caption{Patchwise DINOv3 feature-space analysis. From left to right: balanced normal--normal and normal--anomalous patch distances; cross-image rank-1 distance from each anomalous patch to its nearest normal patch; and fraction of normal patches in its cross-image $k=10$ neighborhood.}
	\label{fig:feature_space_analysis}
	\vspace{-3ex}
\end{figure*}

\begin{figure}[t]
	\centering
	\includegraphics[width=\columnwidth]{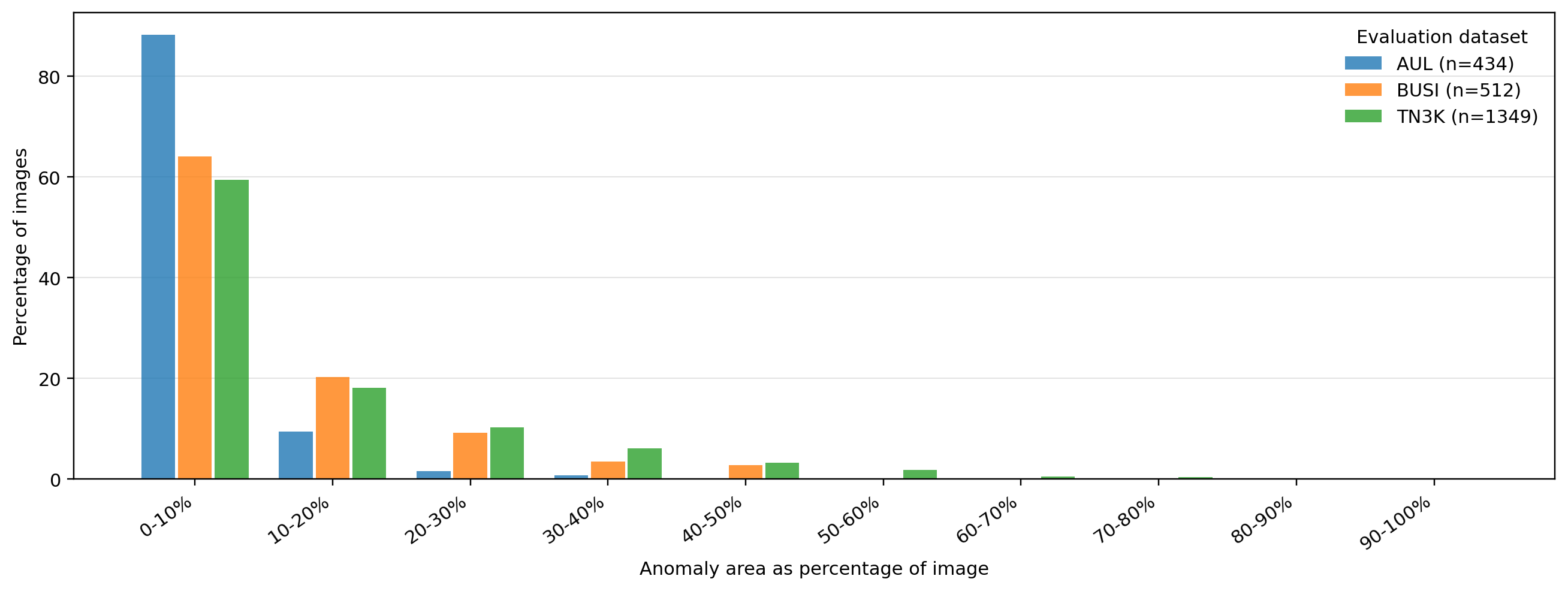}
	\vspace{-2ex}
	\caption{Distribution of the image area covered by annotated pathology in the three evaluation datasets.}
	\label{fig:mask_area_analysis}
	\vspace{-3ex}
\end{figure}

Figure~\ref{fig:feature_space_analysis} summarizes the patchwise feature-space results.
Cross-dataset normal--normal distances remain larger than normal--anomalous distances: their means are $0.805$ versus $0.770$ for the filtered pool and $0.802$ versus $0.761$ for the full pool.
The same pattern occurs within datasets, where normal--normal and normal--anomalous means are $0.738$ and $0.710$ for the filtered pool, and $0.728$ and $0.717$ for the full pool.
For AUL, BUSI, and TN3K, the mean cross-image rank-1 distances to normal patches are $0.265/0.265$, $0.367/0.365$, and $0.414/0.406$ for the filtered/full pools, showing only marginally closer matches in the full pool.
The corresponding mean normal-neighbor fractions for $k=10$ are $0.709/0.722$, $0.419/0.461$, and $0.415/0.540$.
Thus, AUL anomalies mostly overlap normal neighborhoods, whereas BUSI and filtered TN3K more often neighbor anomalous patches across images.
The larger cross-dataset distances indicate substantial source-domain structure.
The normal-neighbor fractions differ between the training pools, most notably for TN3K.
The distributions in Figure~\ref{fig:mask_area_analysis} have median annotated pathology areas of only $2.4\%$, $6.0\%$, and $7.5\%$ for AUL, BUSI, and TN3K, respectively.

Thus, pathology-related neighborhoods exist for some anatomies, but visible pathology does not form a consistent out-of-distribution signal across datasets in the patch-token space.
The small annotated regions further illustrate that models must identify localized pathology despite substantial normal anatomy and source variation.

\section{Discussion}\label{SCM}
Our results show that broad multi-source ultrasound anomaly detection remains challenging: PatchCore variants perform strongest, reaching AUROC values up to $0.82$, $0.82$, and $0.83$ on TN3K, BUSI, and AUL, but their $F1_{\max}$ values remain limited at $0.40$, $0.33$, and $0.18$.
Filtering sources with insufficiently documented pathology status improves dataset semantics but does not make the task uniformly easy.

The reconstruction-based diffusion methods DeCo-Diff and AnoDDPM show limited localization performance, possibly because the heterogeneous training distribution encourages a general ultrasound prior that reconstructs plausible anomalous structures together with normal anatomy.

PatchCore performs better than the reconstruction-based methods, suggesting that feature-space nearest-neighbor comparison may be more robust for this setting.
The DINOv3 backbone gives the highest AUROC on TN3K and AUL, while the WRN backbone gives the strongest AUPRO on TN3K and BUSI. 
We therefore choose PatchCore-DINOv3 for the augmentation experiment because it combines strong ranking behavior with a pretrained patch-token representation that may benefit from additional acquisition-like variability in the memory bank.
The augmented variant improves region overlap in several cases, but its lack of consistent AUROC or $F1_{\max}$ gains suggests that mild appearance and geometric perturbations alone do not solve the separation of normal variation and pathology.
Even with stronger pretrained visual features, absolute localization quality remains limited, indicating that normal ultrasound structures and pathological findings can share local texture, speckle statistics, and boundary appearance; this may also contribute to the weak EfficientAD results.
The patchwise DINOv3 analysis supports this interpretation: BUSI and TN3K show lower normal-neighbor fractions than AUL, while larger cross-dataset distances indicate that feature geometry reflects both pathology and source domain without producing consistent pathology outliers.

\ac{SADUSI} should therefore be understood not only as a dataset for training, but also as a \textbf{stress test for the generalization ability of self-supervised anomaly detection methods in medical ultrasound}.
The goal is not to claim that all pathologies can be detected from single images, but to highlight the gap between current methods and the desired ability to model broad normal ultrasound appearance while still assigning high anomaly scores to visible structural findings.
Important limitations remain, including incomplete coverage of anatomical regions, uneven source sizes, heterogeneous annotation quality, and the limited separability of normal tissue and visible pathology in the analyzed patch-feature space.

\section{Conclusion}\label{AAA}
We introduced \ac{SADUSI}, a multi-source ultrasound dataset for self-supervised anomaly detection across diverse anatomical regions and acquisition settings.
Our experiments show that current methods struggle to separate visible structural anomalies from heterogeneous normal ultrasound appearance.
Feature-memory methods outperform reconstruction-based diffusion methods, but even the strongest PatchCore variants remain limited in pixel-level localization.
The patchwise feature-space analysis further suggests that visible pathology is often not cleanly separated from normal tissue and source-domain variation in pretrained representations.
These findings motivate methods that learn broad ultrasound normality without losing sensitivity to localized structural abnormalities.

\newpage
\bibliographystyle{IEEEtran}
\bibliography{paper}

\end{document}